\pdfoutput=1
\documentclass[11pt]{article}
\usepackage []{EMNLP2023}
\usepackage{algorithm}
\usepackage{algorithmic}

\usepackage{times}
\usepackage{latexsym}
\usepackage{multirow}
\usepackage{xcolor}
\usepackage{arydshln}
\usepackage{amssymb}
\usepackage{textcomp}
\usepackage{booktabs}

\usepackage{latexsym}
\usepackage{comment}

\usepackage{graphicx}
\usepackage{float}
\usepackage{wrapfig}

\usepackage{longtable}
\usepackage{comment}
\usepackage{appendix}
\usepackage{microtype}
\usepackage{inconsolata}

\usepackage{color, colortbl}

\title{ Automatic Restoration of Diacritics for Speech Data Sets}
\author{Sara Shatnawi$^1$,   Sawsan Alqahtani$^2$,  Hanan Aldarmaki$^1$ \\
$^1$Mohamed bin Zayed University of Artificial Intelligence 
\\
$^2$Princess Nourah Bint Abdulrahman University \\
$^1$\texttt{\texttt{\fontsize{10}{12}\selectfont \{sara.shatnawi;hanan.aldarmaki\}@mbzuai.ac.ae}, 
$^2$\texttt{\fontsize{10}{12}\selectfont saalqhtani@pnu.edu.sa} }
}

\begin{document}

\maketitle
\begin{abstract}

Automatic text-based diacritic restoration models generally have high diacritic error rates when applied to speech transcripts as a result of domain and style shifts in spoken language. In this work, we explore the possibility of improving the performance of automatic diacritic restoration when applied to speech data by utilizing parallel spoken utterances. In particular, we use the pre-trained Whisper ASR model fine-tuned on relatively small amounts of diacritized Arabic speech data to produce rough diacritized transcripts for the speech utterances, which we then use as an additional input for diacritic restoration models. The proposed framework consistently improves diacritic restoration performance compared to text-only baselines. Our results highlight the inadequacy of current text-based diacritic restoration models for speech data sets and provide a new baseline for speech-based diacritic restoration.

\end{abstract}

\section{Introduction}

The Arabic script consists of primary alphabetic characters and secondary diacritics. The alphabetic characters represent consonants and long vowels, while diacritics include short vowels, consonant doubling, and \textit{nunation}, which is an additional `n' sound that indicates indefinite nouns in standard and classical Arabic.
Since these diacritics are peripheral rather than main alphabetical characters, proficient Arabic language users typically omit them. As a result, most text resources, including speech transcriptions, are non-diacritized, and currently available Arabic datasets are heavily under-specified for pronunciation. Contextual information provides a basis for implicitly filling out the missing information for proficient speakers, but new learners, nonnative speakers, as well as low-resource speech recognition and synthesis models, often struggle to identify the correct sense and pronunciation due to the lack of diacritics. Consequently, many tasks aimed at achieving higher performance necessitate diacritizing the script as a preliminary step.
To address these challenges, text-based diacritic restoration models (i.e., models that take undiacritized text as input and produce diacritized output) 
have been employed to address the issue of missing diacritics for text and speech applications \cite{fashwan2016rule, fadel2019arabic, fadel-etal-2019-neural,al2020arabic,alkhamissi2020deep, obeid2022camelira}.  
In speech applications, such as Automatic Speech Recognition (ASR) and Text-to-Speech synthesis (TTS), text-based diacritic restoration models have been employed in various ways. Some previous works employed automatic text-based diacritizers to restore the diacritics in speech transcriptions and train a diacritized ASR system \cite{al2014lexical, abed2019diacritics}. 
Alternatively, ASR models can be trained without diacritics, and text post-processing can be employed afterwards to restore the diacritics \cite{aldarmaki2023diacritic}. In TTS, datasets are curated carefully for phonetic coverage, so the training speech transcriptions are carefully annotated and manually diacritized. 
For instance,   
the Classical Arabic Text-To-Speech corpus (ClArTTS) \cite{kulkarni2023clartts} was extracted from a recorded classical Arabic audiobook and was manually diacritized and verified to ensure consistency. The reliance on manual diacritization means that all available datasets for TTS are relatively small, making TTS a low-resource application in Arabic. Automatic diacritization is often used in deployed TTS systems to pre-process the text before synthesis.
  
Uni-modal text-based diacritic restoration models may not be optimal for speech applications \cite{aldarmaki2023diacritic}. Speech utterances are typically less structured than text and may have unusual grammar, repetitions, or missing context, which results in domain and style shifts that lead to poor generalization of these models when applied on speech transcripts. 
Further, in ASR applications, the output may contain transcription errors and misspellings that further sabotage the text diacritization models used in post-processing.   The existence of paired text and speech data presents an opportunity for incorporating an additional modality for disambiguation and diacritic restoration. As shown in \citet{aldarmaki2023diacritic}, ASR models trained to directly produce diacritics outperform text-based diacritizers applied on ASR outputs by a large margin. Given the existence of large speech datasets that contain paired undiacritized texts with speech audios, we explore the potential to improve the performance of automatic diacritic restoration for these datasets.\footnote{As an example, the QASR data set contain $\sim$2000 hours of transcribed Arabic speech data, mostly undiacritized \cite{mubarak2021qasr}.} 
This kind of automatic diacritization could potentially enable the development of large diacritized speech corpora for both ASR and TTS. In particular, we look into whether the speech signal 
could facilitate more robust diacritization for speech-based datasets compared to text-only diacritic restoration models. To that end, we propose a diacritic restoration framework 
that incorporates a pre-trained diacritized speech recognition model. Our experiments show that the framework improves performance compared to an equivalent text-only model, which presents a promising direction for speech-based diacritic restoration. 
Our findings can be summarized as follows:
\begin{enumerate}
\item The proposed diacritic restoration framework results in lower diacritic error rates for read Classical Arabic speech compared to all text-only diacritizers, resulting in a 45\% relative reduction in diacritic error rate compared to the best-performing baseline.  

\item We experiment with both Transformer and LSTM-based architectures with similar scales. While both result in lower error rates compared to the text-only baselines, the LSTM model results in overall better performance. 

\item The performance of the proposed framework partially depends on the performance of the ASR model used to produce the provisional diacritics. Since diacritized datasets are limited for Modern Standard Arabic (MSA) and Dialectal Arabic (DA), more work is needed to improve performance for these variants. 

\end{enumerate}

\section{Related Work}
\subsection{Text-based diacritic restoration} \label{sec:related}
Early approaches for Arabic diacritic restoration mainly relied on morphological rules. For example, \citet{fashwan2016rule} proposed a  rule-based approach to address the case ending diacritization problem in Modern Standard Arabic text. This system relied on morphological and syntactic analyses, taking into account the part-of-speech of each word and its position within the sentence. Morphological analysis has been used as the basis for diacritization in several models, such as MADAMIRA \cite{pasha2014madamira} and the recent Camelira multi-dialectal morphological disambiguator \cite{obeid2022camelira}. In these systems, diacritic restoration is a result of complete morphological analysis and disambiguation, rather than a stand-alone objective.
In recent years, researchers have investigated different neural network-based architectures for stand-alone Arabic diacritization systems. These methods do not rely on morphological analyzers, dictionaries, or feature engineering, but rather use sequence tagging frameworks, leveraging their ability to capture patterns in Arabic text implicitly through end-to-end training. Architectures include feed-forward networks \cite{fadel-etal-2019-neural}, recurrent neural networks (RNNs) \cite{abandah2020accurate,al2020arabic,fadel-etal-2019-neural}, convolutional neural networks (CNNs) \cite{alqahtani-etal-2019-efficient}, and bidirectional LSTM networks possibly followed by Conditional Random Fields (CRF) \cite{al2020arabic}. Some of the most commonly used toolkits and APIs for Arabic diacritization, including Farasa\footnote{\url{farasa.qcri.org/diacritization}}, ALI\_Soft\footnote{\url{ali-soft.com}},  Shakkala\footnote{\url{github.com/Barqawiz/Shakkala}},  Mishkal\footnote{\url{github.com/linuxscout/mishkal}}, and Camelira\footnote{\url{camelira.abudhabi.nyu.edu}}, do not clearly disclose model details or training data, but are often used in practical applications for their convenience.

\subsection{Speech-based diacritic restoration}
To the best of our knowledge, the use of speech data in automatic diacritization has rarely been explored in previous research. The work most closely related to ours in terms of problem formulation is \citet{vergyri2004automatic}, where they explore the use of acoustic and morphological information to automatically restore diacritics in dialectal Arabic speech. The proposed approach employs the EM algorithm to automatically optimize the best diacritic combination using either a morphological analyzer for generating all possible diacritics in an utterance, or using all possible diacritization without morphological constraints. 
The resulting diacritizations were used to construct a word pronunciation network for the acoustic model. 
The model achieved roughly 11\% and 23\% DER with and without morphological analysis, respectively, on the Egyptian CallHome corpus that contains diacritized transcripts in a romanized form. 
Using diacritics to train speech recognition or synthesis systems is desirable, but this requires manually diacritized training data to learn accurate mappings between acoustics and vowels. \citet{aldarmaki2023diacritic} demonstrated through controlled experiments that speech recognition models trained with manually diacritized data sets result in much higher diacritic recognition accuracy compared to text-based diacritic restoration models that are used before training (by diacritizing training speech transcripts) or after inference. This study underscores the importance of optimizing diacritic restoration performance for speech data as the text-based models were shown to have poor generalization in the speech domain.

\begin{figure}
\hspace{-1pt}
    \includegraphics [scale=0.5]{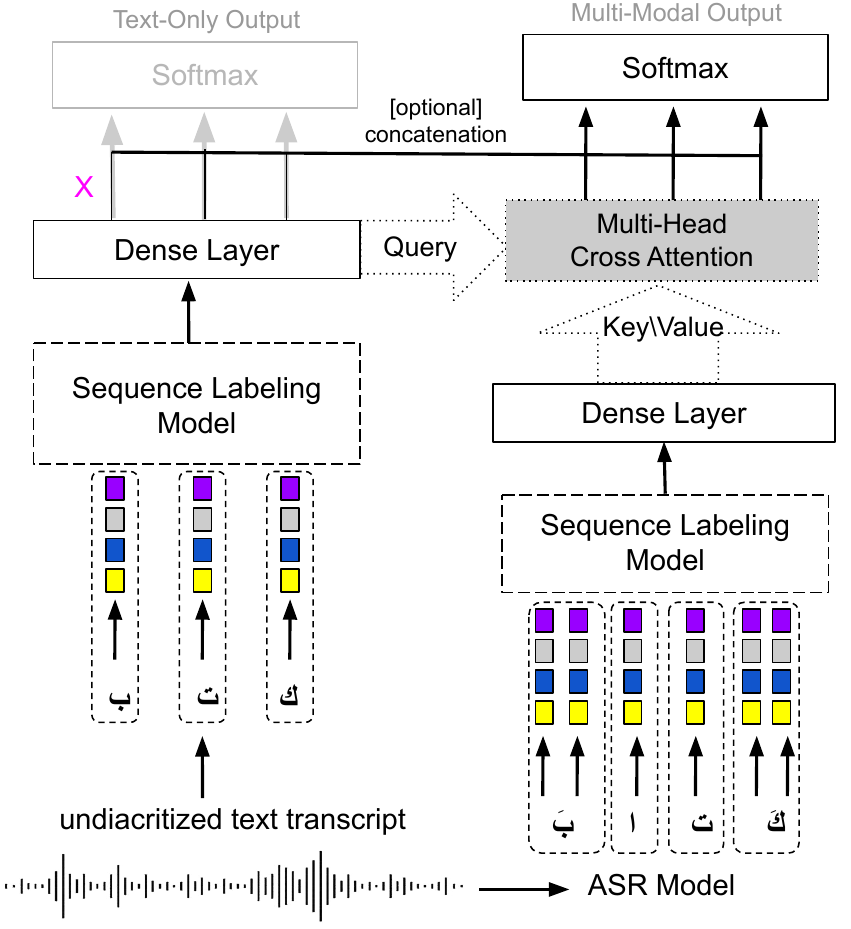} 
    \caption{The proposed diacritic restoration model takes speech utterances and their undiacritized transcripts as input, and produces diacritized text. \textbf{Left}: text-only diacritic restoration, which can be any sequence labeling model. \textbf{Full figure:} Proposed framework, which includes a speech recognition model pre-trained to produce diacritized hypotheses, and a cross-attention mechanism to fuse the two modalities. }
    \label{fig:Text+ASR}
\end{figure}

\section{Proposed Framework}
\label{model_details}
Diacritic restoration is a sequence labeling task:  the input is a sequence of Arabic characters without diacritics, and the output is the target diacritic for each input character, or `no diacritic' if there are none. In our problem setting, we have a dataset that consists of speech utterances along with their undiacritized transcripts. 
Rather than applying a text-based diacritizer on the text transcripts in isolation, we propose a  diacritic restoration framework that incorporates both speech utterances and their text transcripts to produce more accurate diacritics. The proposed model is illustrated in Figure \ref{fig:Text+ASR}. For the speech modality, we utilize a pre-trained ASR model 
to produce provisional diacritized transcripts \footnote{`provisional' because they contain ASR errors that distort both consonants and vowels.}. The undiacritized text transcript and ASR hypothesis are fed into two separate sequence encoders of identical configuration.

To fuse information from the text and speech modalities, we apply cross-attention at the final layer as follows: we use the outputs of the final dense layer of the text encoder (corresponding to the undiacritized text on the left side in Figure \ref{fig:Text+ASR}) as query vectors, and the outputs from the speech side as key and value vectors. Note that ASR predictions are longer than the raw text due to the presence of diacritics and other ASR errors, but the cross-attention mechanism ensures that the final output matches the length of the original undiacritized text. The outputs of cross-attention can be used directly as inputs to the final linear layer with softmax activation for sequence labeling. Note that in this configuration, the prediction relies heavily on the representations from the speech side which contributes the value vectors. To increase the contribution from the text modality, the output of cross-attention can be concatenated with the output of the text encoder (denoted as \textcolor{magenta}{X}  in the figure).

\noindent \paragraph{Notation:} For the rest of the paper, we will refer to this proposed framework as \texttt{Text+ASR};  diacritic restoration models that only rely on the undiacritized text transcripts (as is the case for all the existing baselines) will be referred to as \texttt{Text-Only}. 

\begin{figure*}
 \centering
    \includegraphics [scale=0.5]{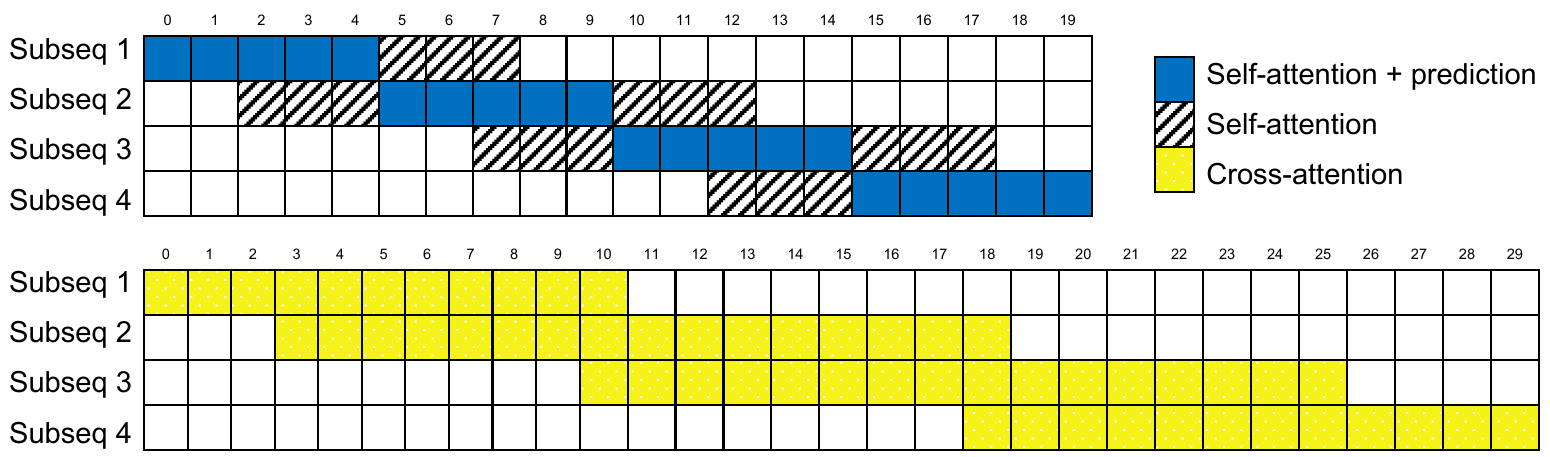}    \caption{\textbf{Top}: Basic transformer self attention and prediction regions with sliding window mechanism for inference. \textbf{Bottom:} Cross-attention region from ASR prediction used in each sub-sequence. }
    \label{fig:attention}
\end{figure*}

\subsection{Sequence Encoder Architecture}
\label{encoders}
The proposed \texttt{Text+ASR} model is conceptually agnostic to the backbone architecture used for speech and text encoders. In this work, we investigated two different architectures to evaluate the efficacy of different sequential models on the proposed approach: Transformer \cite{DBLP:journals/corr/VaswaniSPUJGKP17} and bidirectional LSTM \cite{hochreiter1997long,schuster1997bidirectional}. Note that we use the same architecture for both encoders. 

\noindent \paragraph{Transformer Model:} We utilize the Transformer encoder architecture with position encoding and multi-head self-attention to enable contextual integration across the whole sentence. We employ absolute position embeddings with learned parameters that are optimized along with the rest of the network. We use multiple transformer blocks with multi-head self-attention, followed by a dense layer before the softmax sequence classification.
\noindent \paragraph{bi-LSTM Model:} We use the same architecture described in \citet{fadel-etal-2019-neural}, which achieves the best diacritic restoration performance on our test set. The model consists of an embedding layer followed by several bidirectional LSTM layers. To prevent overfitting, dropout layers are inserted after each bidirectional LSTM layer. The model employs two layers of time-distributed dense units with ReLU activation before the final output layer.

\subsection{Sliding Window Inference}
\label{sliding_window_describtion}
The trained Transformer model with absolute position embeddings can handle sequences up to the maximum length used for training. This kind of position encoding has a poor generalization to longer sequence lengths.\footnote{Most types of absolute and relative positional embeddings have poor length generalization, as shown in \cite{kazemnejad2024impact}.} Furthermore, given the nature of the task, attention to surrounding characters, as well as corresponding regions in the ASR prediction side, is more useful than global attention across two long sequences. Given these reasons, we employ a sliding window mechanism to handle long sequences at inference time and avoid any potential generalization issues. This mechanism is described concretely in Algorithm \ref{alg:slidingwindow} and illustrated in Figure \ref{fig:attention} with $window =5$ and $buffer =3$. Note that this is conceptually similar to the sliding window attention employed in the Longformer model \cite{beltagy2020longformer} but we also proportionally handle cross-attention with the longer ASR sequence, and we apply the method only for inference. 
For the \texttt{Text-Only} model, we employ the sliding window inference, but without the ASR part (i.e. only the top part of Figure \ref{fig:attention}). 

\begin{algorithm}[ht!]
\fontsize{10}{13.6}\selectfont
\caption{Sliding Window Inference}
\label{alg:slidingwindow}

\begin{algorithmic}[1] 
    \REQUIRE $model$, $origText$, $asrText$, $window$,  $buffer$ 
    \ENSURE  $predictions$ 
    \STATE $len \leftarrow \emptyset$ 
    \STATE $len \leftarrow length(origText)$
    \STATE $r \leftarrow length(asrText) / len$ 
    \STATE $start \leftarrow 0$ 
    \WHILE{$start < len$}
        \STATE $safeStart \leftarrow \max(start - buffer, 0)$ 
        \STATE $end \leftarrow start+window+buffer$
        \STATE $end \leftarrow \min(end, len)$ 
        \STATE $s1 \leftarrow origText[safeStart:end]$
        \STATE $s2 \leftarrow asrText[safeStart*r:end*r]$
        \STATE $result \leftarrow model.predict(s1, s2)$ 
        \STATE $end \leftarrow max(start + window, len)$
        \STATE $predictions.insert(result[start:end])$ 
        \STATE $start \leftarrow end $ 
    \ENDWHILE

    \RETURN $predictions$
\end{algorithmic}
\end{algorithm}

\section{Experimental Settings}
\subsection{Datasets}
\label{data_description}
A commonly-used corpus is the cleaned \textbf{Tashkeela} Corpus; this dataset comprises a large collection of Arabic texts mainly from Classical Arabic (CA) literature and religious texts, and a smaller collection in the Modern Standard Arabic (MSA) variety. It consists of 2.3M words spread over 55K lines, which is extracted from the \textbf{original Tashkeela} \cite{zerrouki2017tashkeela} corpus, which has a total of 75M words. We refer to the latter as Tashkeela (Original) but do not use it for training. No speech audios are available in this dataset so we use it for experiments relevant to the \texttt{Text-Only} models. Diacritized speech data sets are rather scarce. We mainly use the CLassical Arabic Text-To-Speech Corpus \textbf{CLArTTS} \cite{kulkarni2023clartts},  which was developed for the purpose of text-to-speech synthesis, so the text transcriptions have been manually diacritized and verified.  The corpus includes approximately 12 hours of recorded speech from a single male speaker (10K relatively short utterances), and about 30 minutes held-out for testing. Since the corpus consists of read Classical Arabic speech, it is consistent with the content of the Tashkeela corpus to enable fair comparison with text-based models.

\subsection{Model Setup}
For the Transformer architecture, we tuned the following hyper-parameters on the ClArTTS validation set: number of transformer blocks, number of attention heads, embedding size, and dropout rate. The reported results are obtained using 128-dimensional token and position embeddings, and 2 Transformer blocks with a feed-forward layer of size 128. We used 4 heads for multi-head attention and 0.2 dropout rate. 
For the LSTM model, we used 128-dimensional vectors for all layers, which include an embedding layer, two bi-LSTM layers, and two dense layers before the final output. Dropout rate of 0.5 was applied after each bi-LSTM. This is similar to the model used in the Shakkelha model \cite{fadel-etal-2019-neural}, which is the strongest baseline on the ClArTTS test set. 
For optimization, we used the Adam optimizer with categorical cross-entropy loss.  
For sliding-window inference, we used a window size 50 and a buffer of 25 tokens on each side
.\footnote{The code for running our experiments is available at \url{https://github.com/SaraShatnawi/Diacritization.git}}

\noindent All models were implemented using the Keras Python library and trained on one Nvidia A100 GPU with 40GB memory. The total number of parameters is $\sim$700K for the \texttt{Text-Only} Transformer and LSTM models and $\sim$1.5M for the \texttt{Text+ASR} variants.

\subsection{ASR Model}
For the ASR module used in the \texttt{Text+ASR} framework, we fine-tuned Whisper \cite{radford2023robust},\footnote{\url{huggingface.com/openai/whisper-medium}} using the training set of the ClArTTS corpus with diacritics. An analysis of this model is provided in \citet{aldarmaki2023diacritic}, but due to apparent inconsistency in their use of ClArTTS train/test splits compared to the official dataset splits,\footnote{We obtained the splits from \url{www.clartts.com}.} we fine-tuned the model from scratch to avoid data leakage.\footnote{The fine-tuned model is available at \href{https://github.com/SaraShatnawi/Diacritization.git}{https://github.com/SaraShatnawi/Diacritization.git}}

\subsection{Baselines}
We compare our proposed \texttt{Text+ASR} model to the \texttt{Text-Only}  baselines that use comparable or similar training dataset. In addition to the Transformer and LSTM models we train, we use popular diacritic restoration models such as the Shakkala model \cite{shakkala}, which was trained with the original Tashkeela corpus of 75M words. Our implementation of the LSTM \texttt{Text-Only} model follows the Shakkelha model described in \citet{fadel-etal-2019-neural}. Shakkelha also includes a variant trained with extra data of 22M words extracted from the original Tashkeela corpus and the Holy Quran. We use it as another baseline that is trained with much more data. For a comprehensive evaluation, we also include results from other popular diacritization APIs even though their training details and datasets are not disclosed, but are worth including due to their popularity: Mishkal \cite{zerrouki2020adawat}, ALI-Soft \cite{alisoft}, Farasa \cite{Farasa}, and Camelira \cite{obeid2022camelira}.

\definecolor{LightCyan}{rgb}{0.88,1,1}

\definecolor{LightYellow}{rgb}{0.8,0.9,0.9}
\definecolor{Gray}{gray}{0.9}
\definecolor{LighterYellow}{rgb}{0.9, 0.9, 0.8}

\section{Results \& Analysis}\label{sec:res}

\begin{table*}[ht!]
\centering
\resizebox{1\textwidth}{!}{%
\begin{tabular}{|c|c|c|c|c|c|}
\hline
\multirow{2}{*}{\textbf{Model}} &
\multirow{2}{*}{\textbf{Train Set}} &
\multicolumn{2}{|c|}{\textbf{Including `no diacritic'}} &
\multicolumn{2}{|c|}{\textbf{Excluding `no diacritic'}} \\
\cline{3-6}
 & & \textbf{w. case ending} & \textbf{w.o case ending} & \textbf{w. case ending} & \textbf{w.o case ending} \\  
  \hline


    \hline 
    \rowcolor{Gray}
  Mishkal \cite{zerrouki2020adawat} & - &
  21.79 &      	        	       	       
  15.80 &
  24.70 &
  17.09 \\
  \hline
  \rowcolor{Gray}      	              	       	
  ali-soft.com & - &
  50.74 &
  47.96 &
  60.03 &
  55.62 \\
  \hline
  \rowcolor{Gray}
  Farasa \cite{Farasa} & - &              	      	  
  19.17 &
  22.05 &
  22.64 &
  26.53 \\
  \hline
  \rowcolor{Gray}
  Camelira \cite{obeid2022camelira} & - &
  29.16 &
  27.36 &
  33.67 &
  32.29 \\
  \hline 
  \rowcolor{LighterYellow}
  Shakkala \cite{shakkala} & Tashkeela (Original)$\dagger$ &
  6.85 &
  5.48 &
  8.02 &
  6.63 \\
  \hline
  \rowcolor{LighterYellow}
  Shakkelha  & Tashkeela & 
  6.02 &
  4.87 &
  6.89 &
  5.70 \\
  \rowcolor{LighterYellow}
  \cite{fadel-etal-2019-neural} & Tashkeela + Extra$\ddagger$&
  4.87 &
  3.54 &
  5.93 &
  4.32 \\
  \specialrule{1pt}{0pt}{0pt}

  \multirow{3}{*}{Text-Only Transformer}   &
  Tashkeela &

  11.73 &
  10.00 &
  13.22 &
  12.39 \\
  &
  CLArTTS &
 20.42   &
   16.96   &
   24.70   &
   20.27 \\
  &       	   		
  Tashkeela + CLArTTS &       
  9.63  &  		
  7.38  &
  11.61 &
  8.91\\
  \cdashline{1-6} 
  \multirow{2}{*}{Text+ASR Transformer}  &
  CLArTTS &
   5.66  &  		
  4.52  &
  6.64 &
  5.45\\
  & 		
  Tashkeela + ClArTTS &
  \textbf{3.63} &
  \textbf{2.71} &
  \textbf{4.07} &
  \textbf{3.17}  \\ 
  \specialrule{1pt}{0pt}{0pt}

  \multirow{3}{*}{Text-Only LSTM}   & 
  Tashkeela &
   6.10   &  		
  4.74   &
 6.97 &
  5.55 \\
  &
  CLArTTS &
  7.97   &  		
  6.60   &
  9.58 &
  7.97  \\
  &       	   		
  Tashkeela + CLArTTS &             	         
  4.93  &  		
  3.55   &
  5.86 &
  4.30\\
    \cdashline{1-6}
  \multirow{2}{*}{ Text+ASR LSTM}  &
  CLArTTS &
   	       	
   2.97  &  		
   2.18  &
  3.03 &
   2.30 \\
  & 		
  Tashkeela + ClArTTS &
	       	       	 
  \underline{\textbf{2.70}} &
  \underline{\textbf{1.83}} &
  \underline{\textbf{ 2.85}} &
  \underline{\textbf{1.99}} \\ 
  \hline
\end{tabular}%
}
\caption{Diacritic Error Rate (DER) \%  for \texttt{Text+ASR} model and the \texttt{Text-Only} models. Baselines that did not disclose details about their training data are shown in Grey background. Baselines that are trained on variants of the Tashkeela corpus are shown in Yellow. $\dagger$ refers to the original Tashkeela corpus of 75M words. $\ddagger$ extra data of 22M words. Tashkeela refers to the subset of 2.3M words from \cite{fadel2019arabic}, which is the one we use for training our models, in addition to the ClArTTS train set. \textbf{Bold} scores refer to the models that have the lowest DER within comparable models with the same architecture. \underline{Underlined} scores refer to the lowest DER overall.}
\label{tab:DER_results_clartt}
\end{table*}

Table \ref{tab:DER_results_clartt} shows the results for our proposed model compared to the baselines using Diacritic Error Rates (DER). We report performance including and excluding the `no diacritic' tag, with and without case ending diacritics.

\paragraph{Performance of previous \texttt{Text-Only} baselines:}
As seen from Table \ref{tab:DER_results_clartt} (grey rows), high error rates are attained by the text-based APIs: Mishakal, ALI-Soft, Farasa, and Camelira on our test set.   
Baselines shown in yellow in Table  \ref{tab:DER_results_clartt}, which are trained on variants of the Tashkeela corpus, achieved the lowest DER overall. 
As previously mentioned, we trained LSTM \texttt{Text-Only} model following the same model and training data as Shakkelha (Tashkeela) to perform our experiments and obtained comparable performance (compared to our \texttt{Text-Only} LSTM model trained on Tashkeela). Shakkelha also has a variant trained with 22M extra words (Tashkeela + Extra) which achieves better performance, demonstrating the effect of dataset size and quality on performance.

\paragraph{Performance of proposed models:}
Our proposed \texttt{Text+ASR} framework results in the best performance overall across all metrics, reducing absolute DER by a large margin compared to their equivalents \texttt{Text-Only} models and compared to all baselines.  The \texttt{Text+ASR} model trained on ClArTTS only, which is a much smaller data set compared to Tashkeela, outperformed most other text-based baselines; and outperformed the best performing Shakkelha model with extra training data when using LSTM (and approached the performance when using Transformer). 
This clearly demonstrates the advantage of using features from speech modality to improve diacritic restoration for speech data. Compared to the \texttt{Text-Only} model trained with ClArTTS alone, the \texttt{Text+ASR} model improved performance by more than 15\% absolute DER when using Transformer and 5\% absolute DER when using LSTM. Further, fine-tuning the \texttt{Text-Only} model trained on Tashkeela with ClArTTS dataset (Tashkeela + ClArTTS) significantly boosts its performance by around 8\% using Transformers and 3.4\% using LSTM. 
Using the \texttt{Text+ASR) framework}, using the combined data improves performance by around 2\% in Transformer but provides similar performance in LSTM compared to using CLaRTTS only.  Overall, LSTM shows better performance compared to the Transformer model. The best performing \texttt{Text-Only} model is the LSTM architecture trained with Tashkeela and fine-tuned on ClArTTS train set, which achieves 4.93\% DER including case ending and `no diacritic', while the equivalent \texttt{Text+ASR} model achieves 2.7\% DER, a 45\% relative improvement.

\begin{figure*}
 \centering
    \includegraphics [scale=0.35]{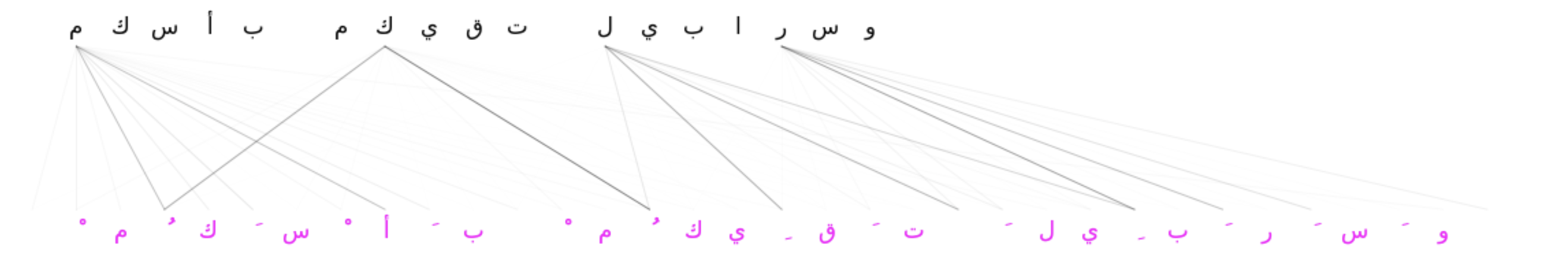}
    \caption{ Cross-attention weights between the undiacritized input (black) and ASR text (magenta). }
    \label{fig:example_attention}
\end{figure*}

\subsection{Analysis}

\paragraph{Text+ASR cross attention layer:}
The cross-attention mechanism plays a major role in integrating information from text and speech inputs to produce the final diacritics. 
Figure \ref{fig:example_attention} shows the cross-attention weights between the raw text and ASR hypothesis obtained from one of the 4 attention heads. We can observe that the weights are higher around the most relevant regions for prediction, regardless of ASR errors. 
\noindent  Table \ref{examples_diac} show an example of a phrase correctly diacritized by the \texttt{Text+ASR} model trained on ClArTTS while the \texttt{Text-Only} models struggle to identify the correct form of the ambiguous phrase.

\begin{table}[ht!]
\centering
\scalebox{0.7}{
\begin{tabular}{|l|c|}
\hline
 \multirow{2}{*}{\textbf{Reference}} &  \multirow{2}{*}{\includegraphics[scale=0.52]{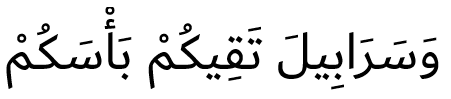}} \\
  &\\
 \hline
  \multirow{2}{*}{\textbf{Camelira }\cite{obeid2022camelira}} &  \multirow{2}{*}{\includegraphics[scale=0.52]{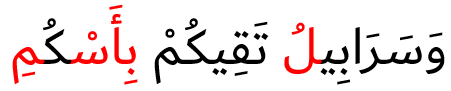}} \\
 &\\
 \hline
  \multirow{2}{*}{\textbf{Shakkala} \cite{shakkala} } &  \multirow{2}{*}{\includegraphics[scale=0.52]{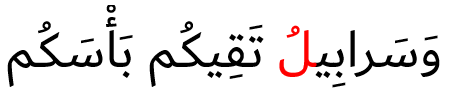}} \\
 &\\
 \hline
 \multirow{2}{*}{\textbf{Text-Only}[ClArTTS] } &  \multirow{2}{*}{\includegraphics[scale=0.6]{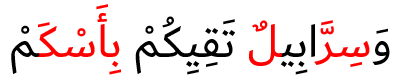}} \\
 &\\
 \hline
 \multirow{2}{*}{\textbf{Text+ASR}[ClArTTS] } &  \multirow{2}{*}{\includegraphics[scale=0.6]{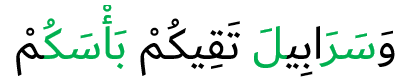}} \\
 &\\
 \hline
\end{tabular}
}
\caption{Diacritization from different models of a phrase from the ClArTTS test set. }
\label{examples_diac}
\end{table}


\begin{table*}
\centering
\resizebox{1\textwidth}{!}{%
\begin{tabular}{|c|c|c|c|c|c|}
\hline
\multirow{2}{*}{\textbf{Model}} &
\multirow{2}{*}{\textbf{Train Set}} &
\multicolumn{2}{|c|}{\textbf{Including `no diacritic'}} &
\multicolumn{2}{|c|}{\textbf{Excluding `no diacritic'}} \\
\cline{3-6}
 & & \textbf{w. case ending} & \textbf{w.o case ending} & \textbf{w. case ending} & \textbf{w.o case ending} \\  
  \hline

\hline
  
  \rowcolor{LightYellow}
  \multicolumn{6}{|l|}{\textbf{Test Set}: Male Speaker from QASR TTS }  \\ 
  
  \hline
  
   \hline
  \rowcolor{Gray}
  Mishkal \cite{zerrouki2020adawat} & - &
  25.24 &      	        	   
  14.47 &
  25.45 &
  12.87   \\
  \hline
  \rowcolor{Gray}      	              	       	
  ali-soft.com & - &
  41.68  &
  36.63 &
  30.90  &
  32.78 \\
  \hline
  \rowcolor{Gray}
  Farasa \cite{Farasa} & - &                	      	       
  28.81 &
  28.81 &
  \underline{21.39} &
  \underline{5.60} \\
  \hline
  \rowcolor{Gray}          	        	       	  
  Camelira \cite{obeid2022camelira} & - &
  30.70  &
  22.64 &
  22.65  &
  12.15  \\
  \hline
   \rowcolor{LighterYellow}      
  Shakkala \cite{shakkala} & Tashkeela (Original)$\dagger$ &
  19.79 &
  11.55 &
  24.12 &
  13.12 \\
  \hline
   \rowcolor{LighterYellow}      
  Shakkelha  & Tashkeela &      	         	       	   
  20.28  &
  12.14 &
  24.83 &
  13.99 \\
   \rowcolor{LighterYellow}       		       	
  \cite{fadel-etal-2019-neural} & Tashkeela + Extra$\ddagger$&
  \underline{19.59} &
  \underline{11.49} &
  23.93&
  13.13 \\
  \hline
  \specialrule{1pt}{0pt}{0pt}
  \multirow{3}{*}{Text-Only Transformer}   &
  Tashkeela &

   23.81  &
  16.06 &
  27.19 &
  16.14 \\
  &
  CLArTTS &
 34.31 &
  34.31 &
  40.91 &
  31.28  \\
  &       	   		
  Tashkeela + CLArTTS &       
   29.28 &
  23.84  &
  33.68 &
  25.59\\
  \cdashline{1-6} 
  \multirow{2}{*}{Text+ASR Transformer}  &
  CLArTTS &
  24.25 &
  16.56  &
  28.22 &
  17.20\\
  & 		
  Tashkeela + ClArTTS &
 \textbf{21.69} &       	        
 \textbf{14.44} &
  \textbf{24.55} &
  \textbf{14.16}  \\ 
  \specialrule{1pt}{0pt}{0pt}

  \multirow{3}{*}{Text-Only LSTM}   & 
  Tashkeela &
 20.20 &
 \textbf{12.02} &
 22.59 &
 \textbf{10.95} \\
  &
  CLArTTS &
   24.93 &
   17.54 &
   28.93 &
   18.15  \\
  &       	   		
  Tashkeela + CLArTTS &             	         
 20.67 &
 12.63 &
 23.49 &
 12.04\\
    \cdashline{1-6}
  \multirow{2}{*}{ Text+ASR LSTM}  &
  CLArTTS &
   	       	
   21.25 &
   14.13 &
   22.95 &
   12.76 \\
  & 		
  Tashkeela + ClArTTS &
	                 	 
 \textbf{19.82} &
12.34 &
\textbf{21.98} &
  11.24 \\ 
  \hline

  \hline
  
  \rowcolor{LightYellow}
  \multicolumn{6}{|l|}{\textbf{Test Set}: Female Speaker from QASR TTS } \\  
  \hline
  
   \hline
  \rowcolor{Gray}       	        	       	   
  Mishkal \cite{zerrouki2020adawat} & - &
  36.28  &      	        	       	       
  27.30 &
  42.62 &
  32.51  \\
  \hline
  \rowcolor{Gray}      	              	       	      	        	         
  ali-soft.com & - &
 42.80   &
 38.39 &
 43.98 &
 45.93 \\
  \hline
  \rowcolor{Gray}    
  Farasa \cite{Farasa} & - &     
  \underline{31.10} &
  \underline{22.29}&
  30.14&
  23.69\\
  \hline
  \rowcolor{Gray}       	        	         Camelira \cite{obeid2022camelira} & - &
  32.40   &
  23.15  &
  34.72 &
  24.86 \\
  \hline
   \rowcolor{LighterYellow}      
  Shakkala \cite{shakkala} & Tashkeela (Original)$\dagger$ &
  34.65 &
  28.63 &
  43.21 &
  35.14  \\
  \hline
   \rowcolor{LighterYellow}      
  Shakkelha  & Tashkeela &  	        	        	       	 
  34.99 &
  29.04 &
  43.56  &
  35.56 \\
       \rowcolor{LighterYellow}      	        	       	
  \cite{fadel-etal-2019-neural} & Tashkeela + Extra$\ddagger$&
  34.01 &
  27.86 &
  42.39 &
  34.15 \\

  \specialrule{1pt}{0pt}{0pt}
  \multirow{3}{*}{Text-Only Transformer}   &
  Tashkeela &

  38.10    &  
  32.69   &     
  35.27   &   	
  21.77 \\
  &
  CLArTTS &
 46.48    &  
  42.56   &     	
  47.96    &   	
  37.20  \\
  &       	   		
  Tashkeela + CLArTTS &       
   39.19 &      	       
  35.35 &       	 
  37.27 &
  26.19\\
  \cdashline{1-6} 
  \multirow{2}{*}{Text+ASR Transformer}  &
  CLArTTS &
    38.59 &    	
  33.28 &   	
  36.62 &  	
  23.17\\
  & 		
  Tashkeela + ClArTTS &
  \textbf{35.69}&
  \textbf{30.72} &
  \textbf{31.84} &
  \textbf{18.85} \\ 
  \specialrule{1pt}{0pt}{0pt}

  \multirow{3}{*}{Text-Only LSTM}   & 
  Tashkeela &
  35.09 &
  \textbf{29.09} &
  30.60&
  16.01 \\
  &
  CLArTTS &
 38.26 &
 33.25 &
 36.04 &
 22.82 \\
  &       	   		
  Tashkeela + CLArTTS &             	         
  35.33&
  29.53&
  31.45 &
  17.05\\
    \cdashline{1-6}
  \multirow{2}{*}{ Text+ASR LSTM}  &
  CLArTTS &
   	       	
 35.59&
 30.69&
 30.25 &
 17.37\\
  & 		
  Tashkeela + ClArTTS &
	       	       	 
 \textbf{34.06} &
 29.11 &
 \underline{\textbf{29.12 }}&
 \underline{\textbf{15.82}} \\ 
  \hline

  \hline

\end{tabular}%
}
\caption{Diacritic Error Rate (DER) \%  of QASR (Male and Female) dataset for Text+ASR (text + ASR) and the text-only models. Baselines that did not disclose details about their training data are shown in Grey background. Baselines that are trained on variants of the Tashkeela corpus are shown in Yellow. $\dagger$ refers to the original Tashkeela corpus of 75M words. $\ddagger$ extra data of 22M words. Tashkeela referes to the subset of 2.3M words from \cite{fadel2019arabic}, which is the one we use for training our models, in addition to the ClArTTS train set. \textbf{Bold} scores refer to the models that have the lowest DER within comparable models with the same architecture. \underline{Underlined} scores refer to the lowest DER overall.}
\label{tab:QASR_DER_Results}
\end{table*}

\subsection{Ablation Experiments}
\paragraph{Impact of concatenation at the last layer:}
We experimented with or without concatenating the outputs of cross-attention layer with the outputs of the \texttt{Text-Only} encoder. As shown in Table \ref{tab:DER_results_conn}, the effect of concatenation varies depending on the encoder architecture. For Transformer, relying solely on the cross-attention outputs provides slightly better performance. On the other hand, the bi-LSTM shows significantly better results with the concatenation. We surmise that this is due to the overall better features obtained from the \texttt{Text-Only} model using bi-LSTM compared to Transformer models. 

\begin{table*}[]
\centering
\resizebox{0.8\textwidth}{!}{%
\begin{tabular}{|c|c|cc|cc|}
\cline{3-6}
\multicolumn{2}{c}{} &
  \multicolumn{2}{|c|}{\textbf{Including `no diacritic'}} &
  \multicolumn{2}{c|}{\textbf{Excluding `no diacritic'}} \\ \hline
\textbf{Model} &
  \textbf{Concatenation} &
  \multicolumn{1}{c|}{\textbf{w. case ending}} &
  \textbf{w.o case ending} &
  \multicolumn{1}{c|}{\textbf{w. case ending}} &
  \textbf{w.o case ending} \\ \hline
Transformer & \checkmark & \multicolumn{1}{c|}{3.83}          & 2.78          & \multicolumn{1}{c|}{4.19}          &\textbf{3.10}          \\ \cdashline{1-6}
Transformer & \texttimes & \multicolumn{1}{c|}{\textbf{3.63}} & \textbf{2.71} & \multicolumn{1}{c|}{\textbf{4.07}} & 3.17 \\ \hline
LSTM        & \checkmark & \multicolumn{1}{c|}{\textbf{2.70}} & \textbf{1.83} & \multicolumn{1}{c|}{\textbf{2.85}} & \textbf{1.99} \\ \cdashline{1-6}
LSTM        & \texttimes & \multicolumn{1}{c|}{7.00}          & 5.96          & \multicolumn{1}{c|}{8.06}          & 7.05          \\ \hline
\end{tabular}%
}
\caption{Diacritic Error Rate (DER) scores on test set with and without concatenation at the last layer. Results are shown for the Text+ASR model trained on Tashkeela + ClArTTS, but the same trend is observed for all variants.}
\label{tab:DER_results_conn}
\end{table*}

\paragraph{Impact of the sliding window for inference:} 
In Section \ref{sliding_window_describtion}, we described a sliding window approach to handle longer sequence lengths at inference. We evaluated the performance both before and after applying this approach. Table \ref{tab:DER_W_WO_Infer_clar} shows notable improvement in results following the application of the proposed inference method.  

\begin{table*}[]
\centering
\resizebox{0.8\textwidth}{!}{%
\begin{tabular}{|c|c|cc|cc|}
\cline{3-6}
\multicolumn{2}{c}{} &
  \multicolumn{2}{|c|}{\textbf{Including `no diacritic'}} &
  \multicolumn{2}{|c|}{\textbf{Excluding `no diacritic'}} \\ \hline
\textbf{Model} &
  \textbf{SW Inference} &
  \multicolumn{1}{c|}{\textbf{w. case ending}} &
  \textbf{w.o case ending} &
  \multicolumn{1}{c|}{\textbf{w. case ending}} &
  \textbf{w.o case ending} \\ \hline
\multicolumn{1}{|c|}{Transformer}                       & \checkmark & \multicolumn{1}{c|}{\textbf{3.63}} & \textbf{2.71} & \multicolumn{1}{c|}{\textbf{4.07}} & \textbf{3.17} \\ \cdashline{1-6}
\multicolumn{1}{|c|}{Transformer} & \texttimes & \multicolumn{1}{c|}{4.12}          & 3.11          & \multicolumn{1}{c|}{4.61}          & 3.54          \\ \hline
\multicolumn{1}{|c|}{LSTM} &
  \checkmark &
  \multicolumn{1}{c|}{\textbf{2.70}} &
  \textbf{1.83} &
  \multicolumn{1}{c|}{\textbf{2.85}} &
  \textbf{1.99} \\ \cdashline{1-6}
\multicolumn{1}{|c|}{LSTM}        & \texttimes & \multicolumn{1}{c|}{2.88}          & 2.07          & \multicolumn{1}{c|}{2.94}          & 2.17          \\ \hline
\end{tabular}%
}

\caption{Diacritic Error Rate (DER) scores on test set with and without applying sliding window inference. Results are shown for the Text+ASR model trained on Tashkeela + ClArTTS, but the same trend is observed for all variants.}
\label{tab:DER_W_WO_Infer_clar}
\end{table*}

\subsection{Additional Experiments on MSA}
As discussed in Section \ref{data_description}, ClArTTS dataset consists of an audiobook in Classical Arabic, and the Tashkeela corpus is derived mostly from Classical Arabic texts. We performed additional experiments to inspect the generalization of the diacritic restoration models trained on these data sets to other variants of Arabic and other speech genres. Diacritized datasets in other Arabic variants are rather scarce, but a small manually diacritic dataset derived from broadcast news has recently been released \cite{baali2023unsupervised}. This dataset (dubbed QASR TTS)\footnote{\url{arabicspeech.org/qasr_tts}} was manually diacritized for TTS and contains one hour of speech by a male speaker and one hour of speech by a female speaker. Note that this dataset contains MSA speech (as opposed to classical Arabic in CLArTTS), has a different genre (broadcast news vs. audiobook), and may have inconsistent speaking and pronunciation styles due to the more casual nature of the recordings. 

The results are shown in Table \ref{tab:QASR_DER_Results}. We show the results separately for the male and female speakers. 
We observe that all models, including all \texttt{Text-Only} baselines, perform poorly on this dataset. While the \texttt{Text+ASR} framework results in some reduction in error rates compared to \texttt{Text-Only} models with identical architectures, the improvements are not consistent or substantial enough. Sources of errors include lexical, domain, and style shifts in the text itself as well as ASR errors. As reported in Table \ref{tab:ASR_Error}, the pre-trained ASR model results in much higher error rates on QASR sets compared to our Classical Arabic test set.  

As an illustrative example, Table \ref{examples_out_domain} shows how ASR performance affects the result of the \texttt{Text+ASR} model in complex ways, sometimes leading to better and sometimes worse performance compared to \texttt{Text-Only}, which eventually leads to similar error rates. Furthermore, all diacritic restoration models used as baselines perform poorly on this set. While we do not know the full sources used to trained the diacritization APIs, we know that the majority of content in Tashkeela consists of Classical Arabic text, and MSA text is estimated to be only 1.15\% of the corpus \cite{zerrouki2017tashkeela}. This shift in the Arabic variant, which results in both lexical and stylistic variations, is likely to be the leading cause for the poor generalization of these models in this dataset.

\begin{table}[ht!]
\centering
\scalebox{0.85}{
\begin{tabular}{|l|cc|cc|}
\hline
\multirow{2}{*}{\textbf{Dataset}} & \multicolumn{2}{c|}{Without Diacritics} & \multicolumn{2}{c|}{With Diacritics} \\ \cline{2-5}
                & \multicolumn{1}{c|}{CER}  & \multicolumn{1}{c|}{WER}   & \multicolumn{1}{c|}{CER}   & \multicolumn{1}{c|}{WER}  \\ \hline
ClArTTS Test & \multicolumn{1}{c|}{2.20} & \multicolumn{1}{c|}{8.02}        & \multicolumn{1}{c|}{2.90}  & \multicolumn{1}{c|}{14.43}        \\ \hline
QASR Female     & \multicolumn{1}{c|}{11.6} & \multicolumn{1}{c|}{36.9}     & \multicolumn{1}{c|}{27.5}  & \multicolumn{1}{c|}{87.3}       \\ \hline
QASR Male       & \multicolumn{1}{c|}{11.1} & \multicolumn{1}{c|}{36.4}      & \multicolumn{1}{c|}{21.06} & \multicolumn{1}{c|}{72.4}         \\ \hline
\end{tabular}
}
\caption{ASR Character Error Rate (CER) \%  and Word Error Rate (WER) \% using the Whisper model fine-tuned on ClArTTS training set. }
\label{tab:ASR_Error}
\end{table}
 
\begin{table}[ht!]
\centering
\scalebox{0.73}{
\begin{tabular}{|l|c|}
\hline
 \multirow{2}{*}{\textbf{Reference}} &  \multirow{2}{*}{\includegraphics[scale=0.6]{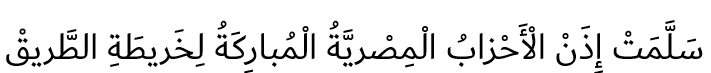}} \\
 &\\
 \hline
 \multirow{2}{*}{\textbf{ASR}} &  \multirow{2}{*}{\includegraphics[scale=0.6]{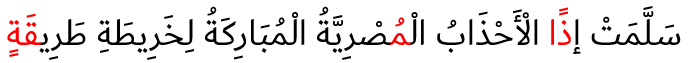}} \\
 &\\

\hline
 \multirow{2}{*}{\textbf{Text-Only}} &  \multirow{2}{*}{\includegraphics[scale=0.6]{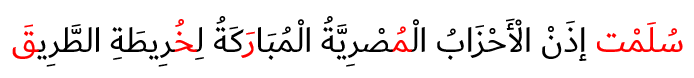}} \\
 &\\
\hline
 \multirow{2}{*}{\textbf{Text+ASR}} &  \multirow{2}{*}{\includegraphics[scale=0.6]{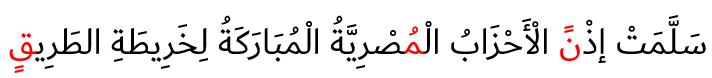}} \\
 &\\
 
 \hline
\end{tabular}}
\caption{Diacritization of a phrase from QASR TTS test set. The diacritic restoration models are our best models trained on both Tashkeela and ClArTTS.}
\label{examples_out_domain}
\end{table}

\section{Discussion \& Conclusion}

We described a framework for diacritic restoration that incorporates input from speech utterances as well as text to improve the performance of diacritic restoration models applied to speech data. The model consists of a pre-trained ASR model that produces provisional diacritized hypotheses, which are incorporated into a sequence labeling model via a cross-attention mechanism. The proposed framework can be applied to any sequence labeling model, and we experimented  with transformer and LSTM architectures.  We also proposed a sliding window inference method that improves the length generalization of the model so it can be applied more robustly to longer sequences. 
The proposed framework consistently improved diacritization performance compared to an equivalent text-only model, leading to significant reductions in diacritic error rates on our test set compared to all existing diacritic restoration models. 
We observed particular performance gains on the Classical Arabic test set which is consistent with the data used for fine-tuning the diacritized ASR  and the diacritic restoration models. The model outperformed all text-based baselines that are trained with much larger text data sets, achieving a 45\% relative reduction in diacritic error rates compared to the best performing baseline. However, given the high variability in speech datasets, additional manually-diacritized datasets are required to enable broader generalization across Arabic variants (such as MSA and dialectal Arabic) and genres (e.g. newswire or casual speech).  
Our results show that highly accurate diacritization can be obtained using a relatively small diacritized speech data set used for training the ASR model and the diacritic restoration models, which can facilitate the production of larger diacritized corpora for speech applications.

\section*{Limitations}
The main limitation of this work is the lack of diacritized speech data sets that could be used for training and testing the framework. While we observed strong performance on the Classical Arabic test set, we could not develop similar models for other variants of Arabic due to limited resources. In addition, we could not evaluate other classical Arabic test sets since we are not aware of any available sets beyond the Quranic data set, which intersects with the Tashkeela corpus. In our analysis, the results appear to be sensitive to ASR performance, which in turn is a factor of the type of training data used to fine-tune the ASR model.

\bibliography{anthology,custom}

\begin{thebibliography}{27}
\expandafter\ifx\csname natexlab\endcsname\relax\def\natexlab#1{#1}\fi

\bibitem[{Abandah and Abdel-Karim(2020)}]{abandah2020accurate}
Gheith Abandah and Asma Abdel-Karim. 2020.
\newblock Accurate and fast recurrent neural network solution for the automatic diacritization of arabic text.
\newblock \emph{Jordanian Journal of Computers and Information Technology}, 6(2).

\bibitem[{Abed et~al.(2019)Abed, Alshayeji, and Sultan}]{abed2019diacritics}
Sa’ed Abed, Mohammad Alshayeji, and Sari Sultan. 2019.
\newblock Diacritics effect on arabic speech recognition.
\newblock \emph{Arabian Journal for Science and Engineering}, 44:9043--9056.

\bibitem[{Al~Hanai and Glass(2014)}]{al2014lexical}
Tuka Al~Hanai and James~R Glass. 2014.
\newblock Lexical modeling for arabic asr: a systematic approach.
\newblock In \emph{INTERSPEECH}, pages 2605--2609.

\bibitem[{Al-Thubaity et~al.(2020)Al-Thubaity, Alkhalifa, Almuhareb, and Alsanie}]{al2020arabic}
Abdulmohsen Al-Thubaity, Atheer Alkhalifa, Abdulrahman Almuhareb, and Waleed Alsanie. 2020.
\newblock Arabic diacritization using bidirectional long short-term memory neural networks with conditional random fields.
\newblock \emph{IEEE Access}, 8:154984--154996.

\bibitem[{Aldarmaki and Ghannam(2023)}]{aldarmaki2023diacritic}
Hanan Aldarmaki and Ahmad Ghannam. 2023.
\newblock Diacritic recognition performance in arabic asr.
\newblock \emph{arXiv preprint arXiv:2302.14022}.

\bibitem[{AlKhamissi et~al.(2020)AlKhamissi, ElNokrashy, and Gabr}]{alkhamissi2020deep}
Badr AlKhamissi, Muhammad~N ElNokrashy, and Mohamed Gabr. 2020.
\newblock Deep diacritization: Efficient hierarchical recurrence for improved arabic diacritization.
\newblock \emph{arXiv preprint arXiv:2011.00538}.

\bibitem[{Alqahtani et~al.(2019)Alqahtani, Mishra, and Diab}]{alqahtani-etal-2019-efficient}
Sawsan Alqahtani, Ajay Mishra, and Mona Diab. 2019.
\newblock \href {https://doi.org/10.18653/v1/D19-1151} {Efficient convolutional neural networks for diacritic restoration}.
\newblock In \emph{Proceedings of the 2019 Conference on Empirical Methods in Natural Language Processing and the 9th International Joint Conference on Natural Language Processing (EMNLP-IJCNLP)}, pages 1442--1448, Hong Kong, China. Association for Computational Linguistics.

\bibitem[{Baali et~al.(2023)Baali, Hayashi, Mubarak, Maiti, Watanabe, El-Hajj, and Ali}]{baali2023unsupervised}
Massa Baali, Tomoki Hayashi, Hamdy Mubarak, Soumi Maiti, Shinji Watanabe, Wassim El-Hajj, and Ahmed Ali. 2023.
\newblock Unsupervised data selection for tts: Using arabic broadcast news as a case study.
\newblock \emph{arXiv preprint arXiv:2301.09099}.

\bibitem[{Barqawi(2017)}]{shakkala}
Zerrouki Barqawi. 2017.
\newblock \href {https://github.com/Barqawiz/Shakkala} {Shakkala, arabic text vocalization}.

\bibitem[{Beltagy et~al.(2020)Beltagy, Peters, and Cohan}]{beltagy2020longformer}
Iz~Beltagy, Matthew~E Peters, and Arman Cohan. 2020.
\newblock Longformer: The long-document transformer.
\newblock \emph{arXiv preprint arXiv:2004.05150}.

\bibitem[{Fadel et~al.(2019{\natexlab{a}})Fadel, Tuffaha, Al-Ayyoub et~al.}]{fadel2019arabic}
Ali Fadel, Ibraheem Tuffaha, Mahmoud Al-Ayyoub, et~al. 2019{\natexlab{a}}.
\newblock Arabic text diacritization using deep neural networks.
\newblock In \emph{2019 2nd international conference on computer applications \& information security (ICCAIS)}, pages 1--7. IEEE.

\bibitem[{Fadel et~al.(2019{\natexlab{b}})Fadel, Tuffaha, Al-Jawarneh, and Al-Ayyoub}]{fadel-etal-2019-neural}
Ali Fadel, Ibraheem Tuffaha, Bara{'} Al-Jawarneh, and Mahmoud Al-Ayyoub. 2019{\natexlab{b}}.
\newblock \href {https://doi.org/10.18653/v1/D19-5229} {Neural {A}rabic text diacritization: State of the art results and a novel approach for machine translation}.
\newblock In \emph{Proceedings of the 6th Workshop on Asian Translation}, pages 215--225, Hong Kong, China. Association for Computational Linguistics.

\bibitem[{Fashwan and Alansary(2016)}]{fashwan2016rule}
Amany Fashwan and Sameh Alansary. 2016.
\newblock A rule based method for adding case ending diacritics for modern standard arabic texts.
\newblock In \emph{16th International Conference on Language Engineering. The Egyptian Society of Language Engineering (ESOLE)}.

\bibitem[{Hochreiter and Schmidhuber(1997)}]{hochreiter1997long}
Sepp Hochreiter and J{\"u}rgen Schmidhuber. 1997.
\newblock Long short-term memory.
\newblock \emph{Neural computation}, 9(8):1735--1780.

\bibitem[{Kazemnejad et~al.(2024)Kazemnejad, Padhi, Natesan~Ramamurthy, Das, and Reddy}]{kazemnejad2024impact}
Amirhossein Kazemnejad, Inkit Padhi, Karthikeyan Natesan~Ramamurthy, Payel Das, and Siva Reddy. 2024.
\newblock The impact of positional encoding on length generalization in transformers.
\newblock \emph{Advances in Neural Information Processing Systems}, 36.

\bibitem[{Kulkarni et~al.(2023)Kulkarni, Kulkarni, Shatnawi, and Aldarmaki}]{kulkarni2023clartts}
Ajinkya Kulkarni, Atharva Kulkarni, Sara Abedalmonem~Mohammad Shatnawi, and Hanan Aldarmaki. 2023.
\newblock Clartts: An open-source classical arabic text-to-speech corpus.
\newblock \emph{arXiv preprint arXiv:2303.00069}.

\bibitem[{Mubarak et~al.(2021)Mubarak, Hussein, Chowdhury, and Ali}]{mubarak2021qasr}
Hamdy Mubarak, Amir Hussein, Shammur~Absar Chowdhury, and Ahmed Ali. 2021.
\newblock Qasr: Qcri aljazeera speech resource a large scale annotated arabic speech corpus.
\newblock In \emph{Proceedings of the 59th Annual Meeting of the Association for Computational Linguistics and the 11th International Joint Conference on Natural Language Processing (Volume 1: Long Papers)}, pages 2274--2285.

\bibitem[{Obeid et~al.(2022)Obeid, Inoue, and Habash}]{obeid2022camelira}
Ossama Obeid, Go~Inoue, and Nizar Habash. 2022.
\newblock Camelira: An arabic multi-dialect morphological disambiguator.
\newblock In \emph{Proceedings of the 2022 Conference on Empirical Methods in Natural Language Processing: System Demonstrations}, pages 319--326.

\bibitem[{Pasha et~al.(2014)Pasha, Al-Badrashiny, Diab, El~Kholy, Eskander, Habash, Pooleery, Rambow, and Roth}]{pasha2014madamira}
Arfath Pasha, Mohamed Al-Badrashiny, Mona~T Diab, Ahmed El~Kholy, Ramy Eskander, Nizar Habash, Manoj Pooleery, Owen Rambow, and Ryan Roth. 2014.
\newblock Madamira: A fast, comprehensive tool for morphological analysis and disambiguation of arabic.
\newblock In \emph{Lrec}, volume~14, pages 1094--1101.

\bibitem[{QCRI(2020)}]{Farasa}
QCRI. 2020.
\newblock \href {https://farasa.qcri.org/diacritization/} {Farasa api diacritization module}.
\newblock Accessed on October 12, 2022.

\bibitem[{Radford et~al.(2023)Radford, Kim, Xu, Brockman, McLeavey, and Sutskever}]{radford2023robust}
Alec Radford, Jong~Wook Kim, Tao Xu, Greg Brockman, Christine McLeavey, and Ilya Sutskever. 2023.
\newblock Robust speech recognition via large-scale weak supervision.
\newblock In \emph{International Conference on Machine Learning}, pages 28492--28518. PMLR.

\bibitem[{Schuster and Paliwal(1997)}]{schuster1997bidirectional}
Mike Schuster and Kuldip~K Paliwal. 1997.
\newblock Bidirectional recurrent neural networks.
\newblock \emph{IEEE transactions on Signal Processing}, 45(11):2673--2681.

\bibitem[{URL(2023)}]{alisoft}
URL. 2023.
\newblock \href {https://www.ali-soft.com} {Ali-soft}.
\newblock Accessed on October 12, 2023.

\bibitem[{Vaswani et~al.(2017)Vaswani, Shazeer, Parmar, Uszkoreit, Jones, Gomez, Kaiser, and Polosukhin}]{DBLP:journals/corr/VaswaniSPUJGKP17}
Ashish Vaswani, Noam Shazeer, Niki Parmar, Jakob Uszkoreit, Llion Jones, Aidan~N. Gomez, Lukasz Kaiser, and Illia Polosukhin. 2017.
\newblock \href {http://arxiv.org/abs/1706.03762} {Attention is all you need}.
\newblock \emph{CoRR}, abs/1706.03762.

\bibitem[{Vergyri and Kirchhoff(2004)}]{vergyri2004automatic}
Dimitra Vergyri and Katrin Kirchhoff. 2004.
\newblock Automatic diacritization of arabic for acoustic modeling in speech recognition.
\newblock In \emph{Proceedings of the workshop on computational approaches to Arabic script-based languages}, pages 66--73.

\bibitem[{Zerrouki(2020)}]{zerrouki2020adawat}
Taha Zerrouki. 2020.
\newblock Towards an open platform for arabic language processing.

\bibitem[{Zerrouki and Balla(2017)}]{zerrouki2017tashkeela}
Taha Zerrouki and Amar Balla. 2017.
\newblock Tashkeela: Novel corpus of arabic vocalized texts, data for auto-diacritization systems.
\newblock \emph{Data in brief}, 11:147--151.

\end{thebibliography}
\bibliographystyle{acl_natbib}

\appendix

\section{Computational Requirements}

We trained our models using one NVIDIA A100-SXM4 GPU with 40GB memory. The GPU processing time for our experiments varies, ranging from 30 minutes to 3 hours, depending on the specific model and dataset employed. Specifically, models utilizing the Tashkeela dataset require approximately three hours for training when implemented with a Transformer architecture and 1 hour and 30 minutes when utilizing LSTMs.
On the other hand, experiments with the CLArTTS dataset exhibit a faster processing time, typically ranging from 40 to 60 minutes for Transformer-based models and half of this time for those employing LSTM. 
\end{document}